\documentclass{article}
\usepackage{iclr2027_conference,times}

\usepackage{amsmath,amsfonts,bm}

\def\eqref#1{equation~\ref{#1}}

\def\1{\bm{1}}

\DeclareMathAlphabet{\mathsfit}{\encodingdefault}{\sfdefault}{m}{sl}
\SetMathAlphabet{\mathsfit}{bold}{\encodingdefault}{\sfdefault}{bx}{n}

\newcommand{\KL}{D_{\mathrm{KL}}}

\usepackage{hyperref}
\usepackage{etoc}
\usepackage{url}
\usepackage{xurl}
\usepackage{doi}
\usepackage{microtype}
\usepackage{enumitem}
\usepackage{booktabs}
\usepackage{tabularx}
\usepackage{pifont}
\usepackage{colortbl}
\usepackage{graphicx}
\usepackage{wrapfig}
\usepackage{needspace}
\usepackage{placeins}
\usepackage{float}
\usepackage{algorithm}
\usepackage{algpseudocode}
\usepackage{flafter}
\usepackage[most]{tcolorbox}
\usepackage{listings}
\definecolor{srdpink}{HTML}{D875A2}
\definecolor{srdpurple}{HTML}{A185C4}
\definecolor{srdgreen}{HTML}{57A887}
\definecolor{srdblue}{HTML}{5F99C7}
\definecolor{srdyellow}{HTML}{D6B44C}

\colorlet{paperink}{black!85}
\colorlet{tableyes}{srdgreen!75!black}
\colorlet{tableno}{srdpink!75!black}
\colorlet{tableaccent}{srdgreen!10!white}
\colorlet{algorithmcomment}{srdblue!75!black}
\colorlet{studentpromptbg}{srdblue!6!white}
\colorlet{studentpromptframe}{srdblue}
\colorlet{studentprompttitle}{srdblue!22!white}
\colorlet{teacherpromptbg}{srdgreen!7!white}
\colorlet{teacherpromptframe}{srdgreen}
\colorlet{teacherprompttitle}{srdgreen!25!white}
\colorlet{peerabg}{srdpink!6!white}
\colorlet{peeraframe}{srdpink}
\colorlet{peeratitle}{srdpink!22!white}
\colorlet{peerbbg}{srdpurple!6!white}
\colorlet{peerbframe}{srdpurple}
\colorlet{peerbtitle}{srdpurple!22!white}
\colorlet{casehighlightbg}{srdyellow!30!white}
\colorlet{casehighlighttext}{srdyellow!40!black}

\newcommand{\cmark}{\textcolor{tableyes}{\ding{51}}}
\newcommand{\xmark}{\textcolor{tableno}{\ding{55}}}
\newtcblisting{studentpromptbox}{
  enhanced,
  breakable,
  listing only,
  colback=studentpromptbg,
  colframe=studentpromptframe,
  colbacktitle=studentprompttitle,
  boxrule=0.6pt,
  arc=1mm,
  title={Student prompt},
  fonttitle=\bfseries,
  coltitle=paperink,
  left=1.4mm,
  right=1.4mm,
  top=1.2mm,
  bottom=1.2mm,
  listing options={
    basicstyle=\ttfamily\small,
    breaklines=true,
    breakatwhitespace=true,
    columns=fullflexible,
    keepspaces=true,
    showstringspaces=false,
    aboveskip=0pt,
    belowskip=0pt
  }
}
\newtcblisting{teacherpromptbox}{
  enhanced,
  breakable,
  listing only,
  colback=teacherpromptbg,
  colframe=teacherpromptframe,
  colbacktitle=teacherprompttitle,
  boxrule=0.6pt,
  arc=1mm,
  title={Teacher prompt},
  fonttitle=\bfseries,
  coltitle=paperink,
  left=1.4mm,
  right=1.4mm,
  top=1.2mm,
  bottom=1.2mm,
  listing options={
    basicstyle=\ttfamily\small,
    breaklines=true,
    breakatwhitespace=true,
    columns=fullflexible,
    keepspaces=true,
    showstringspaces=false,
    aboveskip=0pt,
    belowskip=0pt
  }
}
\hypersetup{hidelinks}

\title{Learning from Unverified Peers via \\ Shared Residual Distillation}
\newcommand{\srdAuthorBlock}{%
  \begin{minipage}[t]{\dimexpr\textwidth-2\tabcolsep\relax}
    \normalfont
    {\fontsize{9.5}{11}\selectfont
      \mbox{%
        \textbf{Feiyang Chen}\textsuperscript{*},
        \textbf{Runze Ma}\textsuperscript{*},
        \textbf{Zichen Wang},
        \textbf{Jiong Lou},
        \textbf{Zihan Wang},
        \textbf{Shuangjia Zheng}\textsuperscript{\ensuremath{\dagger}},
        \textbf{Jie Li}\textsuperscript{\ensuremath{\dagger}}%
      }\par
    }
    \vspace{3pt}
    {\small
      \mbox{Shanghai Jiao Tong University}\quad
      \par
    }
    \vspace{3pt}
    {\footnotesize
      \textsuperscript{*}Equal contribution.\quad
      \textsuperscript{\ensuremath{\dagger}}Corresponding authors.
      \par
    }
  \end{minipage}%
}
\author{\srdAuthorBlock}

\iclrfinalcopy
\newcommand{\method}{Shared Residual Distillation}
\newcommand{\methodabbr}{SRD}
\renewcommand{\KL}{\operatorname{KL}}
\renewcommand{\eqref}[1]{(\ref{#1})}

\newcommand{\stopgrad}{\operatorname{sg}}
\newcommand{\vocab}{\mathcal{V}}

\begin{document}
\etocdepthtag.toc{main}
\maketitle
\fancyhead{}
\fancyhead[L]{Preprint}
\renewcommand{\headrulewidth}{0.4pt}
\begin{abstract}
A language model can generate several solutions to the same problem, some leading to correct answers and others to incorrect ones. Many post-training methods turn these trajectories into learning signals using supervision from reference answers, verifiable feedback, or external teachers. Without such supervision, learning from the model’s own generations becomes less straightforward, as their correctness cannot be reliably assessed. An incorrect solution may nevertheless contain useful intermediate reasoning, so its value as supervision need not be determined solely by its final answer. Can a model improve by learning from its unverified peer solutions without first determining which are correct?
We propose \textbf{\method{} (\methodabbr)}, an on-policy self-distillation method that learns from the model’s own unverified peer trajectories. SRD compares how peers change the prediction distribution relative to a control, retaining adjustments that agree in direction and are conservative in magnitude. This yields supervision without treating any complete solution as correct, and the trained model requires no peers at inference time. Starting from the corresponding base models, \methodabbr{} raises mean AIME24--26 avg@12 from 11.02 to 16.20 on Qwen3-1.7B, from 19.07 to 45.46 on Qwen3-4B, and from 21.39 to 54.44 on Qwen3-8B. Controlled ablations support the value of the shared-residual construction, and we observe no degradation relative to the base models on two general-capability benchmarks.

\end{abstract}
\section{Introduction}
\label{sec:introduction}

Post-training has substantially improved the reasoning capabilities of language models. Reinforcement learning with verifiable rewards (RLVR) uses outcome feedback to form training signals; Group Relative Policy Optimization (GRPO), for example, learns from relative rewards within groups of sampled solutions \citep{deepseekmath}. On-policy distillation (OPD) obtains dense token-level supervision from an external teacher along student-generated trajectories \citep{gkd}. On-policy self-distillation (OPSD) instead enables the model to provide its own supervision by conditioning its teacher on reference answers or solutions available during training \citep{opsd}.

When these supervision sources are unavailable, the model's own generations remain an accessible resource. For the same problem, it can sample reasoning trajectories that follow different strategies and may reach different answers. Test-Time Reinforcement Learning (TTRL) uses majority-vote answers to construct rewards on unlabeled problems \citep{ttrl}. Majority voting thus provides a route to label-free post-training, but still establishes an answer-level reference against which to assign rewards.

\textbf{A trajectory's final answer does not fully describe its value for learning.} Even a solution that ends in an arithmetic or reasoning error can contain useful decompositions, relevant constraints, and valid intermediate derivations. Appendix~\ref{app:case_study} provides a concrete example of useful local guidance from peers with incorrect final answers. We investigate how to extract this local information when reference answers, verifiable feedback, and external teachers are unavailable:

\emph{Can a model learn from its own unverified peer solutions without first determining which answers are correct?}

\Needspace{22\baselineskip}
\begin{wrapfigure}[19]{r}{0.46\textwidth}
  \vspace{-\intextsep}
  \centering
  \setlength{\parskip}{0pt}
  \setlength{\abovecaptionskip}{3pt}
  \includegraphics[width=\linewidth]{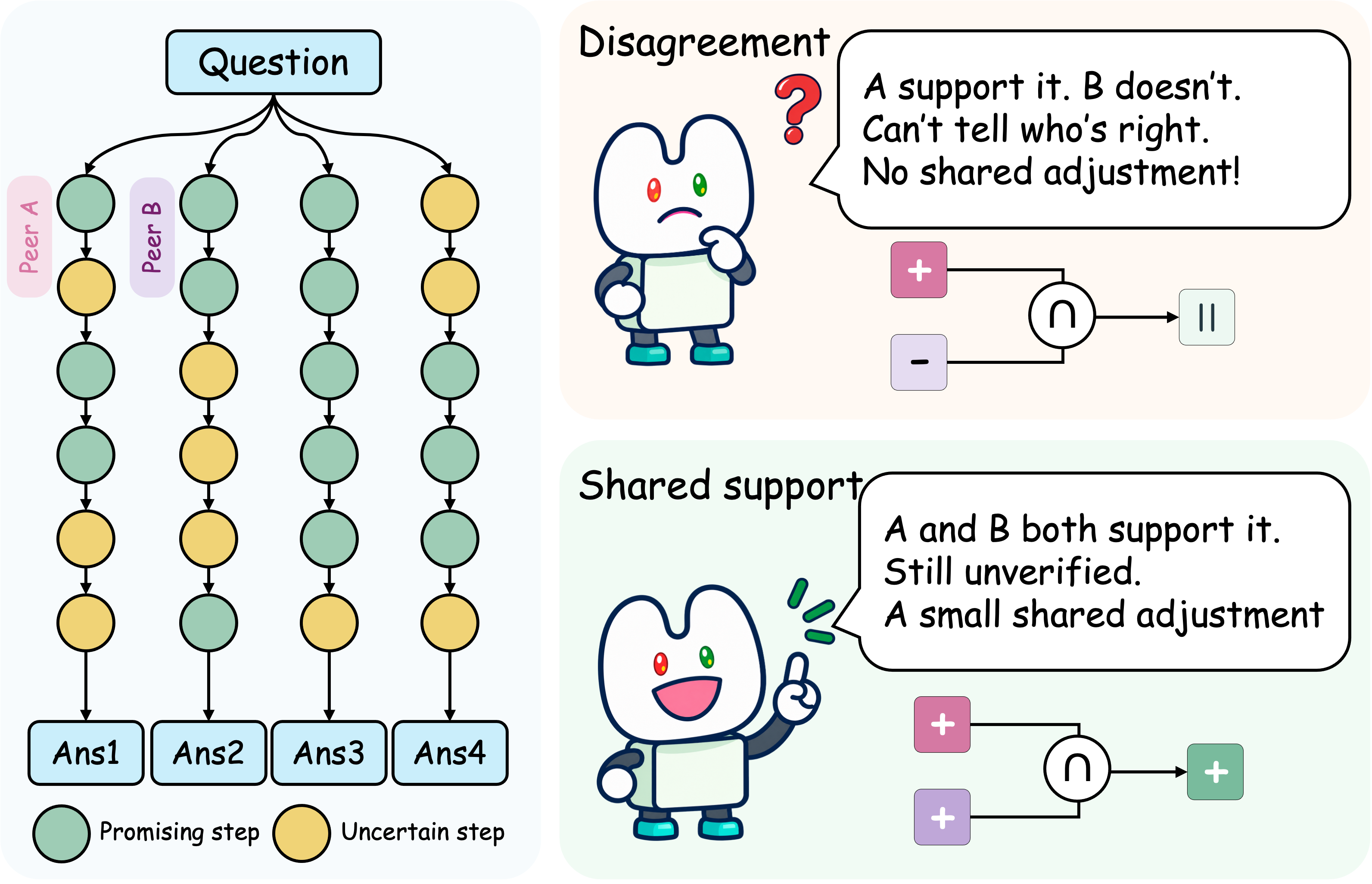}
  \vspace{-0.5\baselineskip}
  \caption{Intuition behind SRD. Sampled solutions may contain useful steps even when their answers are unverified. When peer effects disagree, SRD makes no shared adjustment; when they agree, it retains a conservative adjustment without deciding which peer is correct.}
  \label{fig:peer_intuition}
\end{wrapfigure}

We propose \textbf{\method{} (\methodabbr)}, which turns agreement between peer-induced distributional changes into dense supervision for on-policy self-distillation (Figure~\ref{fig:peer_intuition}). A frozen copy of the initial model evaluates student-generated prefixes while separately conditioned on uniformly sampled same-problem peers and a solution from another problem. The cross-problem solution supplies a common reference for comparing the peers' effects. SRD retains adjustments supported in the \textbf{same direction across peers}, with magnitude bounded by the \textbf{weakest supporting peer}. Applied to the frozen model's distribution under the student input, these adjustments form a teacher target that captures shared local guidance from the sampled solutions. The student learns from this target using only the problem and its own prefix, and requires no peers at inference time.

\par
\ifnum\value{WF@wrappedlines}>1
  \vspace{\dimexpr\value{WF@wrappedlines}\baselineskip-\baselineskip\relax}
\fi
\WFclear

Empirically, \methodabbr{} improves reasoning across model scales. Starting from the corresponding base models, \methodabbr{} raises mean AIME24--26 avg@12 from 11.02 to 16.20 on Qwen3-1.7B, from 19.07 to 45.46 on Qwen3-4B, and from 21.39 to 54.44 on Qwen3-8B. The gains extend to the Harvard--MIT Mathematics Tournament benchmarks. Performance on Instruction-Following Eval (IFEval) and Massive Multitask Language Understanding Pro (MMLU-Pro) is maintained relative to the corresponding base models. Our contributions are summarized as follows:
\begin{itemize}[leftmargin=0pt, labelsep=0.6em, itemindent=0pt, listparindent=0pt, itemsep=4pt, topsep=4pt, parsep=0pt, partopsep=0pt]
    \item We show that unverified model-generated solutions can provide effective post-training supervision without reference answers, correctness rewards, or an external teacher.
    \item We introduce \method{} (\methodabbr{}), which compares token predictions under each peer with those under a cross-problem control. It retains changes that agree in direction across peers and takes the smallest supported magnitude.
    \item We demonstrate mathematical reasoning gains across three Qwen3 model sizes while maintaining performance on the evaluated general-capability benchmarks. Ablations show the value of conservative magnitude selection and the cross-problem residual reference.
\end{itemize}

\section{Related Work}
\label{sec:related}

\suppressfloats[t]
\begin{table}[t]
\centering
\caption{Post-training supervision. Signal is token-level (dense) or outcome-level (sparse); Labels denotes answer-level supervision. --: no distillation teacher.}
\label{tab:method_comparison}
\vspace{3pt}
\small
\setlength{\tabcolsep}{3.5pt}
\renewcommand{\arraystretch}{1.22}
\renewcommand{\tabularxcolumn}[1]{m{#1}}
\begin{tabularx}{\linewidth}{>{\raggedright\arraybackslash}m{82pt}
*{4}{>{\centering\arraybackslash}X}}
\toprule
\textbf{Method}
& \textbf{Samples}
& \textbf{Signal}
& \textbf{Teacher}
& \textbf{Labels} \\
\midrule
RLVR (GRPO) & \cmark\ On-policy & \xmark\ Sparse & -- & Gold \\
External OPD & \cmark\ On-policy & \cmark\ Dense & \xmark\ External & None \\
Reference OPSD & \cmark\ On-policy & \cmark\ Dense & \cmark\ Self & Gold \\
TTRL & \cmark\ On-policy & \xmark\ Sparse & -- & Pseudo \\
\midrule
\rowcolor{tableaccent}
\textbf{SRD (ours)} & \cmark\ On-policy & \cmark\ Dense & \cmark\ Self & \textbf{None} \\
\bottomrule
\end{tabularx}
\end{table}

\paragraph{Reward-based post-training and privileged distillation.}
Post-training learns from human preferences, verifiable rewards, or teacher predictions \citep{instructgpt,dpo,hinton}. GRPO forms relative advantages from rewards within groups of sampled solutions \citep{deepseekmath}; OPD instead uses dense teacher distributions along student-generated prefixes \citep{gkd,minillm}. Privileged information strengthens training-time supervision without changing test-time inputs \citep{lupi,unifying}, and OPSD applies this idea by conditioning a self-teacher on reference solutions \citep{opsd}. CREDIT studies contrastive token credit under environment feedback \citep{credit}. We retain the privileged distillation interface and study target construction when the extra information consists of unverified peer solutions.

\paragraph{Self-generated supervision for post-training.}
Self-Instruct expands instruction data through generation \citep{selfinstruct}, and self-consistency supplies high-confidence reasoning targets for self-improvement \citep{selfimprove}. Self-Rewarding Language Models use model-produced judgments for preference training \citep{selfreward}; TTRL converts answer consensus into rewards on unlabeled problems \citep{ttrl}. Answer consensus also guides self-distillation \citep{coda,ttpo,uopsd}. \methodabbr{} samples unverified peers uniformly and constructs dense supervision from their shared residuals relative to a cross-problem context, without selecting or labeling trajectories by final-answer consensus. Table~\ref{tab:method_comparison} compares supervision sources and granularity.

\paragraph{Label-free inference-time improvement.}
Multiple samples can improve predictions without ground-truth answers at inference. Self-consistency aggregates candidate answers \citep{selfconsistency}, while Tree of Thoughts supports search using model-evaluated intermediate states \citep{tot}. These methods use additional computation to select or construct an answer. They motivate exploiting self-generated reasoning, but do not directly specify a training target. We use multiple solutions to construct local distributional supervision that is transferred into a student generating from the problem alone. Additional related work is provided in Appendix~\ref{app:related}.

\section{Method}
\label{sec:method}

\begin{figure}[!t]
  \centering
  \setlength{\parskip}{0pt}
  \setlength{\abovecaptionskip}{3pt}
  \includegraphics[width=\linewidth,trim=12bp 0 0 0,clip]{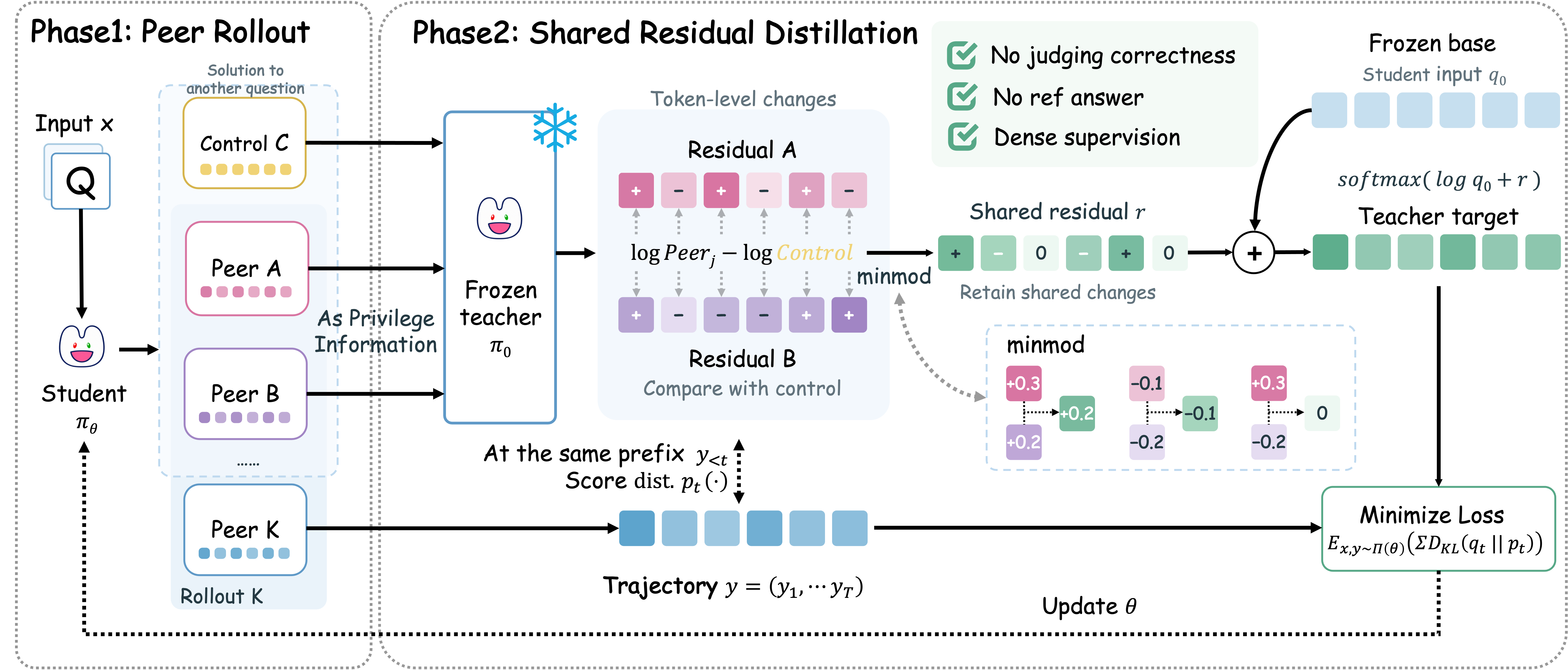}
  \vspace{-0.5\baselineskip}
  \caption{Overview of SRD. \textbf{Phase 1: Peer rollout.} The student samples solutions for each problem, split into peer contexts and target trajectories. \textbf{Phase 2: Shared residual distillation.} A frozen teacher compares peer-conditioned predictions with a cross-problem control and retains conservative shared adjustments. These adjustments modify the frozen model's question-only base distribution to form targets for student updates.}
  \label{fig:method_overview}
\end{figure}

\subsection{Preliminaries}
\label{sec:preliminary}

Let $\mathcal D$ be a dataset of math problems and $\pi_\theta$ a language-model policy. For a problem $x\sim\mathcal D$, let $y=(y_1,\ldots,y_T)\sim\pi_\theta(\cdot\mid x)$ denote a sampled solution. At position $t$, the context $h_t=(x,y_{<t})$ consists of the problem and previously generated tokens; $v\in\vocab$ denotes a candidate next token.

\paragraph{RLVR and GRPO.}
RLVR learns from a verifiable reward $R(x,y)$. GRPO \citep{deepseekmath} samples $G$ trajectories for each problem from $\pi_{\theta_{\mathrm{old}}}$. With $R_i=R(x,y^{(i)})$ and $h_t^{(i)}=(x,y_{<t}^{(i)})$, it computes a group-relative advantage and token-level probability ratio:
\begin{equation}
 A_i=\frac{R_i-\operatorname{mean}_{j}R_j}
 {\operatorname{std}_{j}R_j+\epsilon_0},
 \qquad
 \rho_{i,t}=\frac{\pi_\theta(y_t^{(i)}\mid h_t^{(i)})}
 {\pi_{\theta_{\mathrm{old}}}(y_t^{(i)}\mid h_t^{(i)})},
 \label{eq:grpo}
\end{equation}
where $\epsilon_0>0$ stabilizes the denominator. GRPO maximizes
\begin{equation}
 \mathcal J_{\mathrm{GRPO}}(\theta)
 =\mathbb E\!\left[\frac{1}{G}\sum_{i=1}^G\frac{1}{|y^{(i)}|}\sum_{t=1}^{|y^{(i)}|}
 \left\{\min\!\left[\rho_{i,t}A_i,\,
 \operatorname{clip}(\rho_{i,t},1-\varepsilon,1+\varepsilon)A_i\right]
 -\beta\widehat D_{i,t}\right\}\right],
\label{eq:grpo_objective}
\end{equation}
where $\varepsilon$ controls ratio clipping and $\beta$ weights an estimated Kullback--Leibler (KL) penalty $\widehat D_{i,t}$ against a fixed reference policy. Group normalization avoids a learned value baseline, but still requires rewards to define $A_i$.

\paragraph{OPD and privileged self-distillation.}
OPD \citep{gkd} trains the student using teacher predictions along its own sampled trajectories. Let $\mathcal B$ contain trajectories sampled from the current policy $\pi_{\theta_{\mathrm{old}}}$, with $T_i$ scored positions and prefixes $h_t^{(i)}=(x^{(i)},y_{<t}^{(i)})$. A standard formulation uses the forward KL divergence with a token-mean objective:
\begin{equation}
 \mathcal L_{\mathrm{OPD}}(\theta;q)=
 \mathbb E\!\left[
 \frac{1}{\sum_{i\in\mathcal B}T_i}
 \sum_{i\in\mathcal B}\sum_{t=1}^{T_i}
 \KL\!\left(\stopgrad[q(\cdot\mid h_t^{(i)})]
 \,\middle\|\,\pi_\theta(\cdot\mid h_t^{(i)})\right)
 \right],
 \label{eq:opd}
\end{equation}
where $\stopgrad$ denotes stop-gradient. Trajectories and teacher targets remain fixed during the update.

OPSD \citep{opsd} conditions a frozen self-teacher $\pi_{\bar\theta}$ on a training-only context $z$, such as a reference solution, while leaving the student's input unchanged:
\begin{equation}
 q_z(v\mid h_t)=\pi_{\bar\theta}(v\mid x,z,y_{<t}).
 \label{eq:opsd}
\end{equation}
This follows the privileged-information setting \citep{lupi,unifying}.

\subsection{Shared Residual Distillation}
\label{sec:srd}

\paragraph{Motivation.}
Presenting a solution as context can affect both problem-relevant predictions and generic response patterns. We use a cross-problem solution as a common reference and compare how each same-problem peer changes the teacher's predictions relative to that reference. Agreement determines which residual coordinates are retained, and the weakest supporting peer bounds their magnitude.
\paragraph{Peer contexts.}
For each problem $x\sim\mathcal D$, sample $K>M$ independent trajectories
$Y=\{y^{(1)},\ldots,y^{(K)}\}$ from the current policy. Choose a peer-index set $\mathcal P\subset\{1,\ldots,K\}$ uniformly among all subsets of size $M$, and use the remaining indices
$\mathcal T=\{1,\ldots,K\}\setminus\mathcal P$ as distillation targets.
Thus, each group contributes $M$ peers and $K-M$ held-out target trajectories, and a target trajectory is never its own privileged context.

For the control, randomly sample a trajectory $z_c$ generated for another problem $x_c\neq x$ in the same batch. It reuses an existing rollout and serves as the common reference for all $M$ peer residuals, rather than an additional same-problem peer.

At position $t$ of a target trajectory $i\in\mathcal T$, write $h=(x,y_{<t}^{(i)})$ for its problem and generated prefix. Let $s(x)$ be the student's standard problem prompt and $c(x,z)$ the teacher prompt containing a solution $z$ and guidance on using it. With $z_j=y^{(j)}$, the same frozen initial model gives $M+2$ next-token distributions:
\begin{equation}
\begin{aligned}
 q_0(v\mid h)
   &=\pi_{\bar\theta}(v\mid s(x),y_{<t}^{(i)}),\\
 q_c(v\mid h)
   &=\pi_{\bar\theta}(v\mid c(x,z_c),y_{<t}^{(i)}),\\
 q_j(v\mid h)
   &=\pi_{\bar\theta}(v\mid c(x,z_j),y_{<t}^{(i)}),
   \qquad j\in\mathcal P.
\end{aligned}
\label{eq:views}
\end{equation}
The base $q_0$ uses exactly the student's question-only prompt and target prefix, without additional teacher-side instructions, and is evaluated with frozen initial parameters $\bar\theta$. The current student distribution uses the same input with trainable parameters $\theta$. Each $q_j$ uses a same-problem peer, whereas $q_c$ uses the cross-problem trajectory in the same teacher prompt. All distributions score the current problem $x$ and the same target prefix, without its future continuation; Appendix~\ref{app:prompts} gives the prompts and control substitution.

\paragraph{Shared residual distillation.}
Fix $h$ throughout the following construction and suppress it in the notation. For each same-problem peer, we measure the log-probability change relative to the cross-problem control $q_c$ and subtract its $q_0$-weighted average across the vocabulary:
\begin{equation}
 d_j(v)=\log\frac{q_j(v)}{q_c(v)},\qquad
 \delta_j(v)=d_j(v)-\sum_{u\in\vocab}q_0(u)d_j(u),
 \qquad j\in\mathcal P.
 \label{eq:residual}
\end{equation}
Thus $q_c$ is the residual reference, while $q_0$ supplies the centering weights and the final target's base. Each $\delta_j(v)$ is a scalar centered change for one candidate token; its sign indicates whether the peer-versus-control change lies above or below its $q_0$-weighted average. We compare the residuals \emph{token by token}: when all $M$ are positive or all are negative, retain the smallest magnitude; otherwise retain zero:
\begin{equation}
 r_M(v)=
 \begin{cases}
 \displaystyle\min_{j\in\mathcal P}\delta_j(v),
     & \displaystyle\min_{j\in\mathcal P}\delta_j(v)>0,\\
 \displaystyle\max_{j\in\mathcal P}\delta_j(v),
     & \displaystyle\max_{j\in\mathcal P}\delta_j(v)<0,\\
 0, & \text{otherwise}.
 \end{cases}
 \label{eq:intersection}
\end{equation}
For $M=2$, this is the minmod rule: same sign, smaller magnitude; otherwise zero. At each vocabulary coordinate, the residual magnitude is bounded by the weakest supporting peer-versus-control residual. We apply it to $q_0$, the frozen model's distribution under the student input, to form the teacher target:
\begin{equation}
 q_{\cap}(v)=
 \frac{q_0(v)\exp(r_M(v))}
 {\sum_{u\in\vocab}q_0(u)\exp(r_M(u))}.
 \label{eq:target}
\end{equation}
Repeating this construction at each target prefix provides teacher targets for the question-only student.

\begin{algorithm}[!t]
\caption{Shared Residual Distillation}
\label{alg:srd}
\begin{algorithmic}
\Require Student $\pi_\theta$; frozen model $\pi_{\bar\theta}$; problems $\mathcal D$; $K>M\geq2$.
\Statex \textbf{Notation:} $\mathcal P_x$: $M$ peers for $x$; $\mathcal B$: held-out targets; $\mathcal H(\mathcal B)$: scored target prefixes.
\Statex $\operatorname{Center}_{q_0}(d)=d-\sum_v q_0(v)d(v)$; $\ell_\tau$: clipped loss in Eq.~\eqref{eq:clipped_distillation}.
\For{each iteration}
  \Statex \textcolor{algorithmcomment}{\texttt{// Step 1: sample peer rollouts}}
  \State $\mathcal X\sim\mathcal D,\quad|\mathcal X|\geq2;\qquad Y_x\gets\operatorname{Rollout}(\pi_\theta,x,K),\quad\forall x\in\mathcal X$
  \State $(\{\mathcal P_x\},\mathcal B)\gets\operatorname{UniformSplit}(\{Y_x\},M)$
  \Statex \textcolor{algorithmcomment}{\texttt{// Step 2: construct shared targets}}
  \For{each target $(x,y)\in\mathcal B$}
    \State $z_c\sim\operatorname{Uniform}\!\left(\bigcup_{x'\neq x}Y_{x'}\right)$
    \For{each scored prefix $h$ of $(x,y)$}
      \State $(q_0,q_c,\{q_j\})\gets\operatorname{Score}_{\bar\theta}(h,\mathcal P_x,z_c)$ \hfill Eq.~\eqref{eq:views}
      \State $\delta_j\gets\operatorname{Center}_{q_0}(\log q_j-\log q_c),\quad j\in\mathcal P_x$
      \State $r\gets\operatorname{minmod}(\{\delta_j\}_{j\in\mathcal P_x})$
      \State $q_{\cap}(\cdot\mid h)\gets\stopgrad[\operatorname{softmax}(\log q_0+r)]$
    \EndFor
  \EndFor
  \Statex \textcolor{algorithmcomment}{\texttt{// Step 3: update the student}}
  \State $\mathcal L\gets\operatorname{Mean}_{h\in\mathcal H(\mathcal B)}\ell_\tau\!\left(q_{\cap}(\cdot\mid h),\pi_\theta(\cdot\mid h)\right)$
  \State $\theta\gets\operatorname{OptimizerStep}(\theta,\nabla_\theta\mathcal L)$
\EndFor
\end{algorithmic}
\end{algorithm}

\paragraph{Training objective.}
We inherit OPSD's pointwise-clipped distillation objective \citep{opsd}, replacing its reference-conditioned teacher with $q_{\cap}$. For teacher and student distributions $q$ and $p$, define
\begin{equation}
 \ell_\tau(q,p)
 =\sum_{v\in\vocab}\min\!\left\{
 \bar q(v)\log\frac{\bar q(v)}{p(v)},\,\tau\right\},
 \qquad \bar q=\stopgrad[q].
 \label{eq:clipped_distillation}
\end{equation}
The cap $\tau$ applies to each vocabulary-coordinate contribution before summation; removing it recovers forward KL. Over the batch $\mathcal B$ of held-out target trajectories, we minimize
\begin{equation}
 \mathcal L_{\mathrm{SRD}}(\theta)
 =\mathbb E\!\left[
 \frac{1}{\sum_{i\in\mathcal B}T_i}
 \sum_{i\in\mathcal B}\sum_{t=1}^{T_i}
 \ell_\tau\!\left(q_{\cap}(\cdot\mid h_t^{(i)}),
 \pi_\theta(\cdot\mid h_t^{(i)})\right)
 \right].
 \label{eq:loss}
\end{equation}
Here $T_i$ counts valid target tokens, and the expectation covers rollout, peer, and cross-problem control sampling.

Teacher targets are detached, and the student receives only the problem and its own prefix. Peers, controls, trajectories, and targets remain fixed within an update; fresh groups are sampled from the updated policy. We retain OPSD's clipping threshold and training settings, listed in Appendix~\ref{app:training}. Algorithm~\ref{alg:srd} summarizes the workflow.

The shared target captures agreement relative to a common control at the \emph{current target prefix}, without selecting a complete peer solution as ground truth. At inference, the trained student generates from the problem alone, with no peer sampling or extra teacher calls.

\section{Experiment}
\label{sec:experiments}

Our experiments examine whether shared residuals from unverified peers improve mathematical reasoning, how this supervision compares with reference-based and label-free post-training, and whether the resulting models retain capabilities beyond mathematics.

\subsection{Experiment Setup}
\label{sec:experiment_setup}

We evaluate the Qwen3 family at three model sizes: 1.7B, 4B, and 8B parameters. All main-table \methodabbr{} results use $K=4$ trajectories per problem and $M=2$ peers, leaving two trajectories as distillation targets. A trajectory from another problem in the batch provides the residual control without an additional rollout; Appendix~\ref{app:sensitivity_protocol} varies $K$, rollout length, and $M$.

\paragraph{Datasets and evaluation benchmarks.}
We use the OpenThoughts mathematics subset released with OPSD \citep{opsd}, containing approximately 29.4K problem--solution pairs. \methodabbr{} uses only the problems; reference solutions are available to the reference-based baselines. We evaluate in-domain (ID) mathematical reasoning on AIME24--26, HMMT25/26, and MATH-500 \citep{verify}, and out-of-domain (OOD) retention on IFEval and MMLU-Pro. The OOD benchmarks assess instruction following and broader knowledge-intensive reasoning, respectively. We checked the training set against all evaluation benchmarks and found no overlapping problems. All training and evaluation use non-thinking mode. We report avg@12 on AIME and HMMT, pass@1 on MATH-500, strict prompt-level accuracy on IFEval, and overall accuracy on MMLU-Pro. More details are provided in Appendix~\ref{app:evaluation}.

\paragraph{Baselines.}
We compare with GRPO \citep{deepseekmath} with answer-verified outcome rewards, OPSD \citep{opsd} with reference solutions as privileged teacher context, and TTRL \citep{ttrl} with majority-vote pseudo-rewards. GRPO and TTRL share the same 8192-token rollout and optimization configuration, differing only in whether rewards use verified answers or majority-derived pseudo-labels. OPSD$^\ddagger$ samples two independent target trajectories per problem so that its target batch matches \methodabbr{}; Appendix~\ref{app:opsd_scaling} separately measures the change from the original one-trajectory configuration across all three model sizes. Full settings are provided in Appendix~\ref{app:experiments}.

\begin{table}[!t]
\centering
\caption{Main results across three Qwen3 model sizes (\%; higher is better). AIME and HMMT report avg@12; MATH-500 reports pass@1. Bold marks the best result per model and benchmark. $\dagger$: TTRL trained on the common training split; $\ddagger$: OPSD with two target trajectories per problem.}
\label{tab:main_results}
\vspace{3pt}
\small
\setlength{\tabcolsep}{3.5pt}
\renewcommand{\arraystretch}{1.12}
\begin{tabular*}{\linewidth}{@{\extracolsep{\fill}}l*{8}{c}@{}}
\toprule
& \multicolumn{6}{c}{\textbf{Mathematics (ID)}} & \multicolumn{2}{c}{\textbf{General (OOD)}} \\
\cmidrule(lr){2-7}\cmidrule(l){8-9}
\textbf{Method} & \textbf{AIME24} & \textbf{AIME25} & \textbf{AIME26}
& \textbf{HMMT25} & \textbf{HMMT26} & \textbf{MATH-500}
& \textbf{IFEval} & \textbf{MMLU-Pro} \\
\midrule
\multicolumn{9}{l}{\textit{Qwen3-1.7B}} \\
\addlinespace[2pt]
Base & 13.33 & 8.61 & 11.11 & 5.97 & 8.08 & 74.00 & 66.91 & 43.87 \\
GRPO & 15.28 & 10.83 & 8.89 & 5.69 & 10.86 & \textbf{75.80} & 67.84 & 44.68 \\
TTRL$^\dagger$ & 14.44 & 11.67 & 10.28 & 3.89 & 8.59 & 73.80 & \textbf{68.58} & 44.51 \\
OPSD$^\ddagger$ & 14.17 & 11.39 & 10.56 & 5.97 & 8.33 & 72.60 & 67.10 & 44.17 \\
\midrule
\textbf{\methodabbr{} (ours)} & \textbf{21.11} & \textbf{15.00} & \textbf{12.50} & \textbf{6.11} & \textbf{17.17} & 74.80 & 67.10 & \textbf{45.31} \\
\midrule
\multicolumn{9}{l}{\textit{Qwen3-4B}} \\
\addlinespace[2pt]
Base & 21.11 & 20.83 & 15.28 & 10.56 & 15.91 & 84.20 & 79.30 & 60.61 \\
GRPO & 26.39 & 20.00 & 19.17 & 12.08 & 18.43 & 85.80 & 78.37 & 61.17 \\
TTRL$^\dagger$ & 24.17 & 19.17 & 18.33 & 10.28 & 15.66 & 83.80 & 78.19 & 60.74  \\
OPSD$^\ddagger$ & 22.50 & 26.39 & 25.00 & 10.14 & 18.43 & 83.60 & 79.67 & 60.05 \\
\midrule
\textbf{\methodabbr{} (ours)} & \textbf{47.78} & \textbf{41.11} & \textbf{47.50} & \textbf{25.14} & \textbf{23.48} & \textbf{87.80} & \textbf{80.22} & \textbf{61.52} \\
\midrule
\multicolumn{9}{l}{\textit{Qwen3-8B}} \\
\addlinespace[2pt]
Base & 26.94 & 21.39 & 15.83 & 11.53 & 20.71 & 83.20 & 81.70 & 64.98 \\
GRPO & 27.50 & 23.89 & 17.78 & 12.64 & 18.94 & 85.60 & 80.78 & 65.37 \\
TTRL$^\dagger$ & 24.17 & 25.83 & 17.22 & 12.64 & 17.68 & 87.20 & 80.78 & 65.86 \\
OPSD$^\ddagger$ & 40.56 & 31.94 & 27.50 & 18.06 & 24.24 & 88.20 & 82.26 & 65.98 \\
\midrule
\textbf{\methodabbr{} (ours)} & \textbf{59.72} & \textbf{51.94} & \textbf{51.67} & \textbf{35.97} & \textbf{35.10} & \textbf{93.60} & \textbf{82.81} & \textbf{66.25} \\
\bottomrule
\end{tabular*}
\end{table}

\subsection{Main Results}
\label{sec:main_results}

Table~\ref{tab:main_results} organizes the comparison across three model sizes, mathematical reasoning, and out-of-domain capabilities. The reference-based baselines contextualize performance with access to reference solutions, while TTRL provides a label-free comparison based on inferred answer-level rewards. For all model sizes, we evaluate checkpoints fixed in advance: step 50 for OPSD and \methodabbr{}, and step 300 for GRPO and TTRL. Each method uses the same checkpoint across all benchmarks.

\FloatBarrier

Starting from the corresponding base models, \methodabbr{} raises mean AIME24--26 avg@12 from 11.02 to 16.20 on Qwen3-1.7B, from 19.07 to 45.46 on Qwen3-4B, and from 21.39 to 54.44 on Qwen3-8B. The improvements also extend to HMMT, with no observed degradation relative to the corresponding base models on the two out-of-domain benchmarks. Appendix~\ref{app:pass12} additionally reports pass@12 results on AIME and HMMT.

\begin{figure}[!htbp]
  \centering
  \setlength{\parskip}{0pt}
  \setlength{\abovecaptionskip}{3pt}
  \includegraphics[width=\linewidth]{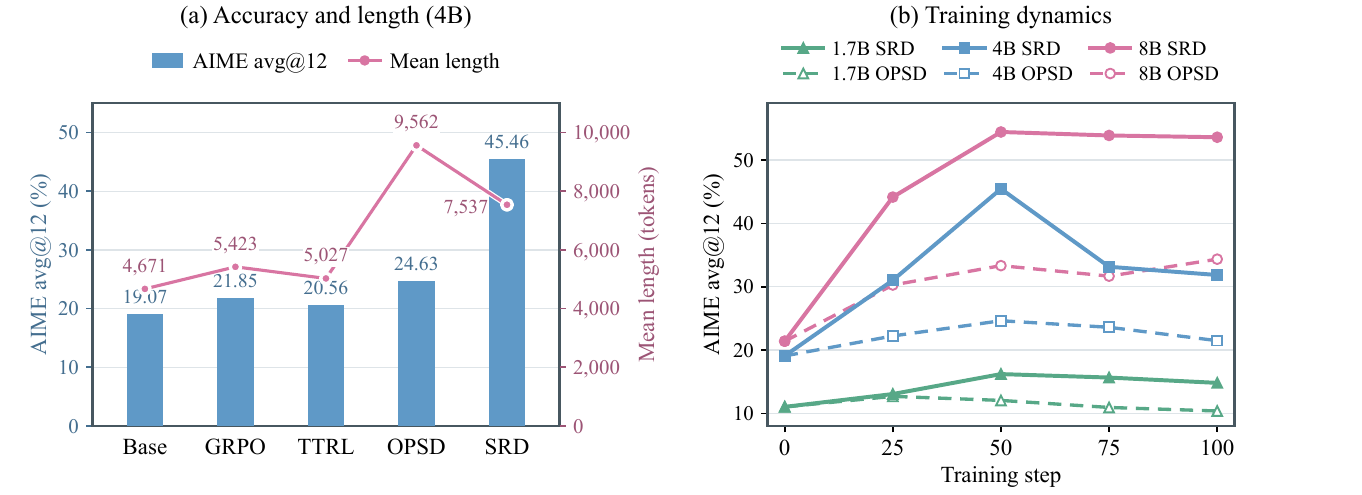}\par
  \caption{AIME accuracy, response length, and training dynamics. (a) Mean AIME24--26 avg@12 (bars) and response length (line) on Qwen3-4B. (b) Mean AIME24--26 avg@12 throughout training for OPSD (dashed) and SRD (solid) at three model sizes.}
  \label{fig:score_length_4b}
\end{figure}

\begin{table}[!t]
\centering
\caption{Qwen3-4B target-construction ablations with clipping at the main-table checkpoint (avg@12, \%). Average is the mean over AIME24--26.}
\label{tab:ablations}
\small
\setlength{\tabcolsep}{6pt}
\renewcommand{\arraystretch}{1.08}
\begin{tabular}{lcccc}
\toprule
\textbf{Variant} & \textbf{AIME24} & \textbf{AIME25} & \textbf{AIME26} & \textbf{Average} \\
\midrule
\multicolumn{5}{l}{\textit{Target construction}} \\
Mean residual & 36.67 & 29.17 & 36.11 & 33.98 \\ %
Random residual mask (matched sparsity) & 41.67 & 32.50 & 41.67 & 38.61 \\
Signed-mask mean & 38.89 & 32.78 & 40.83 & 37.50 \\
Shared residual ($q_0$ reference) & 40.28 & 40.56 & 42.22 & 41.02 \\
\midrule
\rowcolor{tableaccent}
\textbf{Full \methodabbr{} (ours)} & 47.78 & 41.11 & 47.50 & 45.46 \\
\bottomrule
\end{tabular}
\end{table}

Figure~\ref{fig:score_length_4b}(a) compares AIME accuracy and mean response length for the five Qwen3-4B methods at the main-table checkpoints. Mean AIME24--26 avg@12 rises from 19.07 for Base to 45.46 with \methodabbr{}, while mean response length increases from 4,671 to 7,537 tokens. OPSD reaches 24.63 avg@12 with 9,562 tokens on average. \methodabbr{} achieves higher AIME accuracy with shorter responses than OPSD.

Panel~(b) tracks both methods throughout training: \methodabbr{} remains above OPSD at every evaluated post-initialization step across all three model sizes. Both methods show some decline after their initial gains. On Qwen3-4B, mean avg@12 falls from 45.46 to 31.85 for \methodabbr{} and from 24.63 to 21.48 for OPSD between steps 50 and 100. On Qwen3-1.7B, OPSD reaches 10.37 at step 100, below the base model's 11.02. \methodabbr{} on Qwen3-8B retains most of its gains, with mean avg@12 moving only slightly from 54.44 at step 50 to 53.61 at step 100. These patterns indicate sensitivity to model size and training duration in OPSD-based distillation. The stronger retention at 8B suggests that larger models may interpret privileged context more effectively and sustain more stable learning.

\subsection{Ablation Studies}
\label{sec:ablations}

Our central claim is that useful supervision can be extracted from the \emph{shared effects relative to a cross-problem control} of unverified peers, without selecting a preferred solution. We first isolate the target construction on Qwen3-4B, holding the training data, student initialization, rollout budget, target count, peer sampling, and evaluation protocol fixed (Table~\ref{tab:ablations}). We then examine sensitivity to the inherited clipping objective across all three model sizes.

\paragraph{Target construction.}
We first examine how shared direction and conservative magnitude contribute to the teacher target.
The mean-residual variant scores the same four teacher distributions as full \methodabbr{} but replaces the intersection in Eq.~\eqref{eq:intersection} with $(\delta_a+\delta_b)/2$. A matched-sparsity control randomly masks this mean residual to match the nonzero support size of \methodabbr{}, testing whether undirected sparsification is sufficient. Signed-mask mean instead uses the same agreement test as \methodabbr{} to zero sign-disagreeing residual coordinates, but averages the surviving same-sign residuals rather than bounding them by the weakest supporting residual. It isolates the value of conservative magnitude selection under the same sign mask.

Full \methodabbr{} reaches 45.46 mean AIME24--26 avg@12, compared with 37.50 for signed-mask averaging and 38.61 for matched-sparsity random masking. The comparison with signed-mask averaging holds the agreement rule fixed and isolates the contribution of conservative magnitude selection. These results support combining shared direction with a magnitude bounded by the weakest supporting residual.

\paragraph{Residual reference.}
Replacing $q_c$ with $q_0$ in Eq.~\eqref{eq:residual}, while retaining the same centering, shared-residual rule, and $q_0$-based target, yields 41.02 mean AIME24--26 avg@12, compared with 19.07 for the base model and 45.46 for full \methodabbr{}. Shared peer residuals provide substantial improvement with either reference, with the cross-problem control yielding the stronger result.

\paragraph{Effect of pointwise clipping.}
\label{sec:clipping_control}
OPSD introduced pointwise divergence clipping to stabilize distillation \citep{opsd}, and \methodabbr{} inherits this mechanism. Table~\ref{tab:clipping_control} removes clipping from both methods while retaining their respective targets and training settings, with evaluation at the main-table checkpoint. Removing clipping lowers mean AIME24--26 avg@12 for both methods at every model size, consistent with the degradation reported in OPSD's clipping ablation. Both methods therefore achieve their strongest results by combining their respective teacher targets with clipping.

\begin{table}[!ht]
\centering
\caption{OPSD and \methodabbr{} without clipping (\%). Average is the mean over AIME24--26; bold marks column-wise best results within each model.}
\label{tab:clipping_control}
\small
\setlength{\tabcolsep}{3pt}
\renewcommand{\arraystretch}{1.12}
\begin{tabular*}{\linewidth}{@{\extracolsep{\fill}}l*{8}{c}@{}}
\toprule
\textbf{Method}
& \multicolumn{2}{c}{\textbf{AIME24}}
& \multicolumn{2}{c}{\textbf{AIME25}}
& \multicolumn{2}{c}{\textbf{AIME26}}
& \multicolumn{2}{c}{\textbf{Average}} \\
\cmidrule(lr){2-3}\cmidrule(lr){4-5}\cmidrule(lr){6-7}\cmidrule(l){8-9}
& avg@12 & pass@12 & avg@12 & pass@12 & avg@12 & pass@12 & avg@12 & pass@12 \\
\midrule
\multicolumn{9}{l}{\textit{Qwen3-1.7B}} \\
\addlinespace[2pt]
OPSD$^\ddagger$ & 11.39 & 30.00 & 11.11 & 23.33 & 9.17 & \textbf{23.33} & 10.56 & 25.55 \\
\methodabbr{} (ours) & \textbf{12.78} & \textbf{43.33} & \textbf{11.67} & \textbf{30.00} & \textbf{11.67} & 20.00 & \textbf{12.04} & \textbf{31.11} \\
\midrule
\multicolumn{9}{l}{\textit{Qwen3-4B}} \\
\addlinespace[2pt]
OPSD$^\ddagger$ & \textbf{23.61} & 43.33 & 19.72 & 30.00 & 16.11 & 36.67 & 19.81 & 36.67 \\
\methodabbr{} (ours) & \textbf{23.61} & \textbf{50.00} & \textbf{22.22} & \textbf{56.67} & \textbf{20.83} & \textbf{50.00} & \textbf{22.22} & \textbf{52.22} \\
\midrule
\multicolumn{9}{l}{\textit{Qwen3-8B}} \\
\addlinespace[2pt]
OPSD$^\ddagger$ & 29.17 & 53.33 & 20.00 & 33.33 & 18.06 & \textbf{53.33} & 22.41 & 46.66 \\
\methodabbr{} (ours) & \textbf{30.83} & \textbf{56.67} & \textbf{23.61} & \textbf{53.33} & \textbf{19.72} & \textbf{53.33} & \textbf{24.72} & \textbf{54.44} \\
\bottomrule
\end{tabular*}
\end{table}

Under matched clipping conditions, \methodabbr{} has higher mean AIME24--26 avg@12 than OPSD at every model size, both with and without clipping. Even without clipping, \methodabbr{} consistently exceeds both OPSD and the base model on this metric across all three sizes: mean avg@12 rises from the base models' scores of 11.02, 19.07, and 21.39 to 12.04, 22.22, and 24.72 for Qwen3-1.7B, 4B, and 8B, respectively. Against unclipped OPSD, it wins eight of nine individual avg@12 comparisons and ties the remaining one; for pass@12, it wins seven, ties one, and loses one. On AIME25, pass@12 rises from 30.00\% to 56.67\% for Qwen3-4B and from 33.33\% to 53.33\% for Qwen3-8B. These comparisons show that the benefit of shared-residual targets persists when the inherited clipping mechanism is removed.

On Qwen3-4B, the $q_0$-reference variant without clipping reaches 18.61 mean avg@12, slightly below the base model's 19.07. With $q_c$, the score rises from the base model's 19.07 to 22.22. This comparison is consistent with a stabilizing contribution from the cross-problem control when clipping is absent. Appendix~\ref{app:sensitivity_protocol} reports sampling-budget and peer-count sensitivity, including Figure~\ref{fig:sensitivity_panels}.

\section{Conclusion}
\label{sec:conclusion}

We study how a language model can learn from its own unverified solutions without reference answers, correctness labels, or selecting a winning trajectory. We introduce \method{}, which uses independent peer solutions as privileged teacher contexts and constructs local supervision from their shared, conservatively bounded residuals relative to a randomly sampled cross-problem control, applied to the frozen model's distribution under the student input. Starting from the corresponding base models, \methodabbr{} raises mean AIME24--26 avg@12 from 11.02 to 16.20 on Qwen3-1.7B, from 19.07 to 45.46 on Qwen3-4B, and from 21.39 to 54.44 on Qwen3-8B. The improvements extend to HMMT, with no observed degradation relative to the base models on the evaluated out-of-domain benchmarks. These results show that unverified peer generations can provide effective dense post-training supervision even when no trajectory is promoted as a pseudo-label.

\section{Limitations}
\label{sec:limitations}

Peer agreement does not guarantee correct reasoning; \methodabbr{} uses it to identify shared local adjustments without assigning correctness labels to entire solutions. The cross-problem control is intended to reduce generic distributional shifts introduced by additional solution text and make the supervision more specific to the current problem. The extent to which it removes stylistic effects remains unclear, and the control's own content may introduce effects that alter residual directions. Our empirical study focuses on mathematical reasoning, and the effectiveness of \methodabbr{} on open-ended tasks remains untested.

\clearpage
\bibliography{references}
\bibliographystyle{iclr2027_conference}
\clearpage
\appendix
\section{Additional Related Work}
\label{app:related}

\paragraph{Reasoning, verification, and feedback.}
Chain-of-thought prompting exposes intermediate computation \citep{cot,zeroshot}, while outcome rewards and process supervision turn evaluations of reasoning into training signals \citep{deepseekmath,verify,mathshepherd}. The Self-Taught Reasoner (STaR) bootstraps from generated rationales filtered using answer correctness \citep{star}; this differs from learning without correctness labels. Feedback can also be expressed in language: Self-Refine iterates self-feedback \citep{selfrefine}, Reflexion uses verbal feedback across attempts \citep{reflexion}, and tool-interactive critiquing incorporates external checks \citep{critic}. Multiagent debate and Self-Contrast use interaction or contrast among solutions to guide subsequent responses \citep{debate,selfcontrast}. Their effectiveness depends on the model, task, and feedback source \citep{cannotcorrect,surveycorrection}; reinforcement learning can explicitly train correction behavior \citep{score}. These are broader ways to exploit generated reasoning, rather than components of our residual target construction.

\paragraph{Further distillation background.}
Sequence-level distillation transfers generated sequences \citep{seqkd}, while Born Again Networks and Deep Mutual Learning demonstrate useful transfer between similarly parameterized models \citep{bornagain,mutual}. On-policy distillation follows the broader motivation of learning on the learner's state distribution \citep{dagger}. Teacher-generated rationales support reasoning transfer \citep{stepbystep,reasoningteachers,scott}. Weak-to-strong generalization further shows that imperfect supervision need not impose a hard ceiling on a pretrained student's performance \citep{weakstrong}. These results motivate learning from fallible outputs; our focus is how multiple such outputs should influence a single distributional target.

CoDA uses a representative answer-consensus trajectory as privileged context and adds a reference-anchored, KTO-style penalty for minority trajectories \citep{coda}. TTPO conditions on a majority-vote pseudo-answer, distills agreeing rollouts, and penalizes disagreeing rollouts through grouped RL, with token weighting and masking in the respective branches \citep{ttpo}. u-OPSD accepts pseudo-solutions using a self-consistency threshold and distills the resulting teacher on disagreeing trajectories \citep{uopsd}. These approaches use final-answer consensus to construct or route supervision. \methodabbr{} instead samples peers uniformly and checks agreement between their centered residuals at each target prefix and vocabulary coordinate; peers can support the same local adjustment even when their final answers differ.

\paragraph{Consistency and distributional differences.}
Mean Teacher and FixMatch exploit consistency across views, together with labeled supervision, to learn from unlabeled examples \citep{meanteacher,fixmatch}; deep ensembles aggregate predictions from multiple models \citep{ensembles}. Our peers are unverified solutions used as contexts for one frozen model, and we combine their \emph{changes relative to a common cross-problem control}, then apply the shared adjustment to the frozen model's distribution under the student input. Differences between predictive distributions also underlie contrastive decoding, context-aware decoding, and Decoding by Contrasting Layers (DoLa) \citep{contrastivedecoding,contextdecoding,dola}. Those methods steer inference using model, context, or layer contrasts. Related interference-reduction ideas appear in gradient surgery and sign-aware model merging \citep{pcgrad,ties}. \methodabbr{} applies a coordinatewise agreement rule to peer-induced log-probability residuals at student prefixes and uses the resulting teacher target for on-policy distillation. Its specific object is a shared conditional shift, rather than an averaged prediction, a merged parameter vector, or an inference-time decoding rule.

CREDIT studies self-distillation conditioned on environment feedback and subtracts a batch-estimated generic teacher component to refine token-level credit \citep{credit}. Its contrast changes the \emph{problem input} while holding the response prefix and feedback fixed, and the resulting signal enters a reverse-KL update. \methodabbr{} holds the current problem and prefix fixed and changes the \emph{solution context}, comparing each same-problem peer with an unrelated trajectory. It then retains shared centered residuals through minmod and normalizes the adjusted base $q_0$ into a distillation target. Thus our construction combines a context contrast with agreement across peers, using generated solutions without environment feedback.

\clearpage
\section{Case Study: Local Guidance from Incorrect Peers}
\label{app:case_study}

Peers A/B answer incorrectly (correct: 429); C solves another problem. Ellipses mark omissions.

\begingroup
\tcbset{casestudy/.style={
  enhanced, boxrule=0.5pt, arc=1mm,
  left=1.5mm, right=1.5mm, top=1.2mm, bottom=1.2mm,
  before skip=5pt, after skip=5pt,
  fonttitle=\small\bfseries, coltitle=paperink,
  fontupper=\small,
}, casecolumn/.style={
  height=79mm, segmentation style={draw=none}, valign lower=bottom,
  fontupper=\fontsize{9}{10.5}\selectfont\raggedright,
  before upper={
    \setlength{\abovedisplayskip}{3pt}
    \setlength{\belowdisplayskip}{3pt}
    \setlength{\abovedisplayshortskip}{2pt}
    \setlength{\belowdisplayshortskip}{2pt}
  },
}}
\begin{tcolorbox}[casestudy,title={Question},
  colback=studentpromptbg,colframe=studentpromptframe,
  colbacktitle=studentprompttitle]
Adam and Bettie are playing a game. They take turns generating a random number between $0$ and $127$ inclusive. The numbers they generate are scored as follows:
\begin{itemize}[leftmargin=1.2em,nosep,topsep=2pt]
\item If the number is zero, it receives no points.
\item If the number is odd, it receives one more point than the number one less than it.
\item If the number is even, it receives the same score as the number with half its value.
\end{itemize}
If Adam and Bettie both generate one number, the probability that they receive the same score is $\frac{p}{q}$ for relatively prime positive integers $p$ and $q$. Find $p$.
\end{tcolorbox}

\begin{tcbraster}[raster columns=3,raster column skip=2mm,
  raster before skip=0pt,raster after skip=0pt]
\begin{tcolorbox}[casestudy,casecolumn,title={Peer A},
  colback=peerabg,colframe=peeraframe,colbacktitle=peeratitle]
\ldots
Numbers from 0 to 127 (inclusive) are all 7-bit numbers: 0000000 to 1111111 in binary.
\par\smallskip
\ldots
\par\smallskip
Sum of squares:
\[
\begin{aligned}
 &1+49+441+1225\\
 &{}+1225+441+49+1\\
 &{}=\colorbox{casehighlightbg}{\textcolor{casehighlighttext}{\textbf{3912}}}
\end{aligned}
\]
Then:
\[
P=\frac{3912}{128^2}=\frac{3912}{16384}
\]
\ldots
\[
\frac{3912}{16384}=\frac{489}{2048}
\]
\tcblower
\textbf{Final answer:} $\boxed{489}$
\end{tcolorbox}
\begin{tcolorbox}[casestudy,casecolumn,title={Peer B},
  colback=peerbbg,colframe=peerbframe,colbacktitle=peerbtitle]
\ldots
Note that 127 is $2^7-1=1111111_2$, so numbers from 0 to 127 include all 7-bit binary numbers (with leading zeros allowed), \ldots
\[
\sum_{k=0}^{7}\binom{7}{k}^{\!2}=\binom{14}{7}
\]
Compute:
\[
\binom{14}{7}=\cdots=\colorbox{casehighlightbg}{\textcolor{casehighlighttext}{\textbf{12376}}}
\]
\ldots
\[
\frac{12376}{16384}=\frac{1547}{2048}
\]
\tcblower
\textbf{Final answer:} $\boxed{1547}$
\end{tcolorbox}
\begin{tcolorbox}[casestudy,casecolumn,title={Control C (cross-problem)},
  colback=srdyellow!6!white,colframe=srdyellow,
  colbacktitle=srdyellow!22!white]
We are given the equation:
\[
\frac{252}{\Theta}=\underline{3\Theta}+\Theta
\]
\ldots
\[
\underline{3\Theta}=10\cdot3+\Theta=30+\Theta
\]
\ldots
\[
\Theta^2+15\Theta-126=0
\]
\ldots
\[
\Theta=\frac{-15\pm27}{2}
\]
\ldots
\par\smallskip
But $\Theta$ is a digit, so it must be between 0 and 9.
\tcblower
\textbf{Final answer:} $\boxed{6}$
\end{tcolorbox}
\end{tcbraster}

\begin{tcolorbox}[casestudy,title={Student (excerpt)},
  colback=studentpromptbg,colframe=studentpromptframe,
  colbacktitle=studentprompttitle]
We are given a game between Adam and Betty. \ldots\ So the score of a number $n$ is the number of 1s in its binary representation. \ldots\
Each number from 0 to 127 has \colorbox{casehighlightbg}{\textcolor{casehighlighttext}{\textbf{8}}} bits (since $2^8=256$), so we can list how many numbers have a certain number of 1s in their 8-bit binary representation.
\end{tcolorbox}

\begin{center}
\small
\textbf{Next-token probabilities at the highlighted student position.}\\[3pt]
\begin{tabular}{@{}lrr@{}}
\toprule
\textbf{Privileged context / target} & $P(\texttt{7})$ (\%) & $P(\texttt{8})$ (\%) \\
\midrule
Student input ($q_0$) & 18.13 & 79.45 \\
Cross-problem control ($q_c$) & 13.54 & 83.06 \\
Peer A ($q_A$) & 99.96 & 0.01 \\
Peer B ($q_B$) & 99.90 & 0.04 \\
\methodabbr{} ($q_{\cap}$) & 99.83 & 0.06 \\
Reference solution ($q_{\mathrm{ref}}$) & 99.94 & 0.01 \\
\bottomrule
\end{tabular}
\end{center}
\endgroup

The frozen model uses the student's question-only prompt for $q_0$ and the cross-problem solution C as auxiliary context for $q_c$. All rows score the prefix ending in ``Each number from 0 to 127 has'', using full-vocabulary softmax at temperature 1.1. The control does not reverse the token preference: both $q_0$ and $q_c$ favor the incorrect \texttt{8} over \texttt{7}. Both peers retain the correct seven-bit representation despite later arithmetic errors; their shared-residual target assigns 99.83\% to \texttt{7}. In a separate greedy continuation from this prefix, forcing only \texttt{7} and then continuing without peers yields the correct answer 429, versus 6435 without the intervention. This example illustrates that final-answer correctness is not a prerequisite for useful local guidance. Peers with incorrect answers can still provide shared local signals that survive subtraction of the cross-problem control and conservative residual aggregation.

\clearpage
\section{Experimental Details}
\label{app:experiments}

\subsection{Data and Evaluation Protocol}
\label{app:evaluation}

\paragraph{Training data and supervision.}
We use the released OPSD subset, \path{siyanzhao/Openthoughts_math_30k_opsd}.\footnote{\url{https://huggingface.co/datasets/siyanzhao/Openthoughts_math_30k_opsd}}
We use its mathematical problem statements as the common source of training inputs. OPSD supplies the released \texttt{solution} field only to its teacher, and GRPO verifies generated answers against the reference. Neither \methodabbr{} nor the TTRL adaptation uses reference solutions, correctness annotations, or benchmark labels for optimization. Benchmark problems are reserved for evaluation.

\paragraph{Benchmark versions.}
The American Invitational Mathematics Examination (AIME) benchmark combines the first and second examinations of each year, giving 30 problems per annual benchmark. For HMMT25, we evaluate both the February 2025 and November 2025 Harvard--MIT Mathematics Tournament (HMMT) releases and report the arithmetic mean of their two benchmark scores. HMMT26 uses the February 2026 release. We use the MathArena releases for HMMT25/26 and AIME26;\footnote{\url{https://huggingface.co/MathArena}}
the HMMT26 set contains 33 entries and is evaluated in full. MATH-500 uses the 500-problem test subset distributed by HuggingFaceH4.\footnote{\url{https://huggingface.co/datasets/HuggingFaceH4/MATH-500}}
Instruction-Following Eval (IFEval; \citealp{ifeval}) uses the original released prompts and deterministic instruction checkers, while Massive Multitask Language Understanding Pro (MMLU-Pro; \citealp{mmlupro}) uses its full test split and five subject-matched demonstrations from its validation split.\footnote{\url{https://github.com/google-research/google-research/tree/master/instruction_following_eval}; \url{https://github.com/TIGER-AI-Lab/MMLU-Pro}}
ID and OOD are defined relative to the mathematical adaptation domain; IFEval and MMLU-Pro assess general capabilities after mathematical post-training.

\paragraph{Generation and metrics.}
Native chat templates are used with thinking disabled for student sampling, teacher scoring, and evaluation. Mathematical evaluations are zero-shot, with a request to place the final answer in \texttt{\textbackslash boxed\{\}}. We use a generation budget of 32,768 tokens, a context limit of 40,960, temperature 0.7, top-$p$ 0.8, top-$k$ 20, and presence penalty 1.0. Evaluation runs with vLLM 0.11, one complete model per device, and tensor parallelism of one; independent problem shards provide data parallelism. All compared methods use the same evaluator, prompts, and per-benchmark decoding settings.

For $N$ problems, let $c_{ij}$ indicate whether response $j$ to problem $i$ is correct. The reported sampling accuracy is
\begin{equation}
 \operatorname{avg@12}=\frac{100}{12N}\sum_{i=1}^{N}\sum_{j=1}^{12}c_{ij}.
 \label{eq:avg12}
\end{equation}
This is an estimate of single-sample accuracy, not pass@12 or majority-vote accuracy. MATH-500 uses one response per problem under the same mathematical decoding settings, yielding pass@1. Mathematical correctness is evaluated from the last boxed answer with symbolic equivalence checking through Math-Verify.\footnote{\url{https://github.com/huggingface/Math-Verify}}
Missing or unparseable answers count as incorrect. IFEval and MMLU-Pro use one greedily decoded response, with an 8,192-token output budget and no added math-specific instruction. IFEval's primary score requires every verifiable instruction in a prompt to pass the strict checker; MMLU-Pro reports the fraction of test questions with the correct selected option.

\subsection{Training Configurations}
\label{app:training}

\paragraph{Shared distillation setup.}
Within each model size, OPSD$^\ddagger$ and \methodabbr{} use a frozen copy of that model's initialization as teacher and keep the student's prompt question-only. The Qwen3-4B configuration uses Low-Rank Adaptation (LoRA; \citealp{lora}) with rank 64 and scaling factor 128 on the query, key, value, output, gate, up, and down projections. Both methods use a learning rate of $5\times10^{-6}$, a linear schedule over 100 optimizer steps, and maximum gradient norm 0.1. Both use the pointwise-clipped objective defined in Eqs.~\eqref{eq:clipped_distillation}--\eqref{eq:loss}, following OPSD \citep{opsd}. Student and teacher distributions use scoring temperature 1.1, and the coordinatewise upper cap is $\tau=10^{-6}$(1.7B and 4B, while 8B is $\tau=10^{-7}$). Full-vocabulary summation precedes averaging over valid scored tokens. OPSD uses its reference-conditioned teacher target in place of $q_{\cap}$. Table~\ref{tab:distillation_hyperparameters} separates sampling budgets from scored prefixes.

\begin{table}[ht]
\centering
\caption{Training configuration for Qwen3-4B distillation. OPSD$^\ddagger$ matches SRD's number of scored target trajectories without generating unused peer contexts. Both methods use a 100-step schedule.}
\label{tab:distillation_hyperparameters}
\small
\setlength{\tabcolsep}{7pt}
\renewcommand{\arraystretch}{1.12}
\begin{tabular*}{\linewidth}{@{\extracolsep{\fill}}lcc@{}}
\toprule
\textbf{Setting} & \textbf{OPSD$^\ddagger$} & \textbf{\methodabbr{}} \\
\midrule
Problems per update & 32 & 32 \\
Rollouts per problem & 2 & 4 \\
Target trajectories per update & 64 & 64 \\
Maximum rollout length & 1,024 & 1,024 \\
Maximum scored target prefix & 1,024 & 1,024 \\
Maximum training context & 20,000 & 20,000 \\
Rollout temperature / top-$p$ / top-$k$ & 1.1 / 0.95 / 20 & 1.1 / 0.95 / 20 \\
Frozen teacher contexts & Reference solution & None / peers $a,b$ / control $c$ \\
Learning rate & $5\times10^{-6}$ & $5\times10^{-6}$ \\
LoRA rank / scaling & 64 / 128 & 64 / 128 \\
Coordinate clipping threshold & $10^{-6}/10^{-7}$ & $10^{-6}/10^{-7}$ \\
Maximum gradient norm & 0.1 & 0.1 \\
\bottomrule
\end{tabular*}
\end{table}

\paragraph{Rationale for inherited clipping.}
\citet{opsd} report that a small set of stylistic tokens can contribute disproportionately large KL terms relative to math-related tokens, motivating a pointwise cap to prevent these contributions from dominating distillation. Their Section~4.3.3 and Figure~4 show that clipping prevents performance collapse on Qwen3-1.7B. We retain this stabilizer for both OPSD and \methodabbr{}, holding the distillation objective fixed when comparing their teacher targets. Following the official OPSD implementation, both our OPSD reproduction and \methodabbr{} optimize a pointwise-clipped surrogate rather than the unmodified teacher--student KL. Capping positive summands can therefore produce negative training losses in either method, as noted in OPSD's official documentation.\footnote{\url{https://github.com/siyan-zhao/OPSD\#key-opsd-arguments}} Section~\ref{sec:clipping_control} evaluates both methods without the cap and shows that \methodabbr{} retains higher mean AIME24--26 avg@12 than OPSD and the base model at every model size.

\paragraph{Peer sampling and target count.}
\methodabbr{} samples four trajectories per problem, uniformly reserves two distinct trajectories as peer contexts, and scores the remaining two. A control trajectory is randomly drawn from those generated for a different problem in the same batch and is shared across the same-problem peer residuals. Thus, 32 problems still yield 128 sampled trajectories and 64 distillation targets per update. The default construction scores four frozen-teacher distributions per target prefix: $q_0$, $q_a$, $q_b$, and $q_c$. Reusing an in-batch trajectory avoids extra generation, but scoring $q_c$ adds a teacher evaluation. OPSD$^\ddagger$ needs no peer contexts, so it samples and scores two trajectories per problem, yielding the same 64 targets without generating unused trajectories. The default rollout and scored-prefix limits are both 1,024 tokens. The rollout-length ablation extends generation to 2,048 or 4,096 tokens while retaining the 1,024-token scoring limit (Appendix~\ref{app:sensitivity_protocol}).

\subsection{Baseline Details}
\label{app:baseline_details}

Table~\ref{tab:baseline_hyperparameters} consolidates the controlled configurations for GRPO and TTRL. Both use the same LoRA modules, rank, and scaling as the distillation methods, and share their rollout group, decoding, optimizer, and update budget. Their sole algorithmic difference is the reward label: GRPO checks the normalized final answer against the reference, while TTRL checks it against the current group's normalized majority answer. Our TTRL entry applies this construction to the common training split rather than adapting on benchmark inputs, as marked by $\dagger$ in Table~\ref{tab:main_results}. OPSD$^\ddagger$ follows Table~\ref{tab:distillation_hyperparameters}; Appendix~\ref{app:opsd_scaling} reports its scaling control separately.

\begin{table}[ht]
\centering
\caption{Controlled configurations for the GRPO and TTRL baselines, which differ only in their reward labels.}
\label{tab:baseline_hyperparameters}
\footnotesize
\setlength{\tabcolsep}{4pt}
\renewcommand{\arraystretch}{1.10}
\begin{tabularx}{\linewidth}{>{\raggedright\arraybackslash}p{0.27\linewidth}
*{2}{>{\centering\arraybackslash}X}}
\toprule
\textbf{Setting} & \textbf{GRPO} & \textbf{TTRL$^\dagger$} \\
\midrule
Supervision & Verified reward & Majority pseudo-reward \\
Sampling & On-policy & On-policy \\
Problems per update & 32 & 32 \\
Responses per problem & 8 & 8 \\
Trajectories per update & 256 & 256 \\
Maximum response length & 8192 & 8192 \\
Temperature / top-$p$ / top-$k$ & 1.1 / 0.95 / 20 & 1.1 / 0.95 / 20 \\
Optimization & LoRA 64 / 128 & LoRA 64 / 128 \\
Learning rate & $10^{-6}$ & $10^{-6}$ \\
Training budget & 300 updates & 300 updates \\
Warmup updates & 10 & 10 \\
Policy-ratio interval & $[0.8,1.28]$ & $[0.8,1.28]$ \\
Added KL / entropy penalty & None / none & None / none \\
\bottomrule
\end{tabularx}
\end{table}

\subsection{Student and Teacher Prompts}
\label{app:prompts}

The boxes below reproduce the user-message contents used by \methodabbr{} before applying the model's native chat template. Braced fields are replaced by the current problem and a sampled solution trajectory. Thinking is disabled for the student and all teacher evaluations. The student prompt is used for rollout generation and for scoring both the student distribution and $q_0$: their input tokens and attention masks are identical, with $q_0$ evaluated using frozen initial weights. The teacher prompt is used separately for each same-problem distribution $q_j$ and the control $q_c$, scoring the same target prefix. For $q_c$, only the \texttt{\{peer\_solution\}} field is replaced by a trajectory from another problem; the current problem and prompt wording are retained. The shared prompt explicitly states that the auxiliary solution may or may not address the same problem and does not assume its correctness. For the OPSD baseline, we reuse the original reference-conditioned teacher prompt from \citet{opsd}.

\begin{studentpromptbox}
{problem}

Please reason step by step, and put your final answer within \boxed{}.
\end{studentpromptbox}

\begin{teacherpromptbox}
{problem}

An auxiliary solution is provided below. It may or may not address the same problem:
=== Auxiliary Solution Begin ===
{peer_solution}
=== Auxiliary Solution End ===

Use the auxiliary solution as potentially useful privileged information, but do not assume that its reasoning or final answer is correct. Solve the problem using your own independent reasoning, checking any information you use from it. Think step by step, explore different approaches, and don't be afraid to backtrack or reconsider if something doesn't work out:

Please reason step by step, and put your final answer within \boxed{}.
\end{teacherpromptbox}

\clearpage
\subsection{Pass@12 Results}
\label{app:pass12}

Table~\ref{tab:pass12_results} reports pass@12 on AIME24--26 and HMMT25/26 at the corresponding main-table checkpoints. This metric measures the percentage of problems with at least one correct response among twelve samples, complementing the average per-response accuracy reported in the main results. HMMT25 averages the February and November benchmark scores, following Appendix~\ref{app:evaluation}.

\begin{table}[!ht]
\centering
\caption{Pass@12 on AIME and HMMT (\%; higher is better). The best score within each model--benchmark block is bold, including ties. Method annotations follow Table~\ref{tab:main_results}.}
\label{tab:pass12_results}
\small
\setlength{\tabcolsep}{6pt}
\renewcommand{\arraystretch}{1.12}
\begin{tabular*}{\linewidth}{@{\extracolsep{\fill}}l*{5}{c}@{}}
\toprule
\textbf{Method} & \textbf{AIME24} & \textbf{AIME25} & \textbf{AIME26}
& \textbf{HMMT25} & \textbf{HMMT26} \\
\midrule
\multicolumn{6}{l}{\textit{Qwen3-1.7B}} \\
\addlinespace[2pt]
Base & 33.33 & 23.33 & 26.67 & 16.67 & 27.27 \\
GRPO & 40.00 & 23.33 & 20.00 & 18.33 & 24.24 \\
OPSD$^\ddagger$ & 33.33 & 26.67 & 26.67 & \textbf{21.67} & 24.24 \\
TTRL$^\dagger$ & 36.67 & 33.33 & 20.00 & 11.67 & 24.24 \\
\midrule
\textbf{\methodabbr{} (ours)} & \textbf{46.67} & \textbf{40.00} & \textbf{33.33} & 20.00 & \textbf{30.30} \\
\midrule
\multicolumn{6}{l}{\textit{Qwen3-4B}} \\
\addlinespace[2pt]
Base & 46.67 & 36.67 & 46.67 & 26.67 & 39.39 \\
GRPO & 50.00 & 36.67 & 56.67 & 30.00 & 30.30 \\
OPSD$^\ddagger$ & 56.67 & 43.33 & 46.67 & 33.33 & 36.36 \\
TTRL$^\dagger$ & 46.67 & 40.00 & 50.00 & 30.00 & 30.30 \\
\midrule
\textbf{\methodabbr{} (ours)} & \textbf{76.67} & \textbf{70.00} & \textbf{73.33} & \textbf{45.00} & \textbf{42.42} \\
\midrule
\multicolumn{6}{l}{\textit{Qwen3-8B}} \\
\addlinespace[2pt]
Base & 53.33 & 40.00 & 36.67 & 28.33 & 39.39 \\
GRPO & 56.67 & 36.67 & 46.67 & 33.33 & 33.33 \\
OPSD$^\ddagger$ & 70.00 & 50.00 & 56.67 & 41.67 & \textbf{42.42} \\
TTRL$^\dagger$ & 50.00 & 40.00 & 46.67 & 33.33 & 33.33 \\
\midrule
\textbf{\methodabbr{} (ours)} & \textbf{80.00} & \textbf{76.67} & \textbf{80.00} & \textbf{63.33} & \textbf{42.42} \\
\bottomrule
\end{tabular*}
\end{table}

\methodabbr{} achieves the highest pass@12 on all three AIME editions at every model size, showing that its gains extend to problem coverage under a twelve-sample budget.
\FloatBarrier

\clearpage
\section{Ablation Details}
\label{app:ablation_details}

\subsection{Ablation Protocol}
\label{app:ablation_protocol}

The core Qwen3-4B ablations in Table~\ref{tab:ablations} are evaluated at the main-table checkpoint and preserve the full method's problem batch, group size, uniform peer sampling, target count, optimization schedule, and evaluation protocol unless that quantity is ablated.
Mean-residual, matched-sparsity, signed-mask mean, and full \methodabbr{} all score $q_0$, $q_a$, $q_b$, and $q_c$, using the same cross-problem control sampling rule and residuals in Eq.~\eqref{eq:residual}. The $q_0$-reference variant instead sets $d_j(v)=\log(q_j(v)/q_0(v))$, retaining the same $q_0$-weighted centering and shared-residual aggregation; it requires only $q_0$, $q_a$, and $q_b$. Every residual-based target retains $q_0$ as its base. The matched-sparsity variant first forms $(\delta_a+\delta_b)/2$. At every scored position, it then uniformly samples, without replacement, exactly as many vocabulary coordinates as are nonzero under the corresponding \methodabbr{} intersection and zeros the remainder. The random mask is sampled independently of residual sign and magnitude. This matches the per-position support size---and hence the residual zeroing rate---while removing the method's rule for selecting which directions survive. Signed-mask mean instead applies the same all-positive or all-negative test as Eq.~\eqref{eq:intersection}, but assigns a surviving coordinate the arithmetic mean $M^{-1}\sum_{j=1}^{M}\delta_j(v)$. It therefore preserves the sign mask while removing the conservative minimum-magnitude rule.

\subsection{Additional Ablations}
\label{app:additional_ablations}

Table~\ref{tab:additional_ablations} reports secondary target controls at the main-table checkpoint. The prompt-only teacher receives the problem with instructions to reason independently, explore alternative approaches, and backtrack, without any solution trajectory. Its additional teacher-side instructions distinguish it from $q_0$, while the absence of a solution context distinguishes it from $q_c$. This variant tests whether instructions alone can supply useful distillation supervision. Two-peer probability averaging tests whether directly averaging predictions conditioned on the same uniformly sampled peers is sufficient. Section~\ref{app:sensitivity_protocol} reports sensitivity to sampling budgets and peer count.

\begin{table}[!h]
\centering
\caption{Additional Qwen3-4B target-construction ablations. Metrics follow Table~\ref{tab:ablations}; Average is the arithmetic mean over AIME24--26.}
\label{tab:additional_ablations}
\footnotesize
\setlength{\tabcolsep}{6pt}
\renewcommand{\arraystretch}{1.0}
\begin{tabular}{lcccc}
\toprule
\textbf{Variant} & \textbf{AIME24} & \textbf{AIME25} & \textbf{AIME26} & \textbf{Average} \\
\midrule
\multicolumn{5}{l}{\textit{Additional target controls}} \\
Prompt-only teacher & 7.22 & 4.17 & 8.33 & 6.57 \\
Two-peer probability mean & 40.56 & 29.72 & 37.50 & 35.93 \\
\bottomrule
\end{tabular}
\end{table}

The prompt-only teacher uses the following user message:

\begin{teacherpromptbox}
Problem: {problem}

Make sure you truly understand the reasoning behind each step. Using your own words and independent reasoning, solve the problem above. Think step by step, explore different approaches, and don't be afraid to backtrack or reconsider if something doesn't work out:

Please reason step by step, and put your final answer within \boxed{}.
\end{teacherpromptbox}

\subsection{Sampling and Peer-Count Sensitivity}
\label{app:sensitivity_protocol}

\begin{figure}[H]
  \centering
  \includegraphics[width=\textwidth]{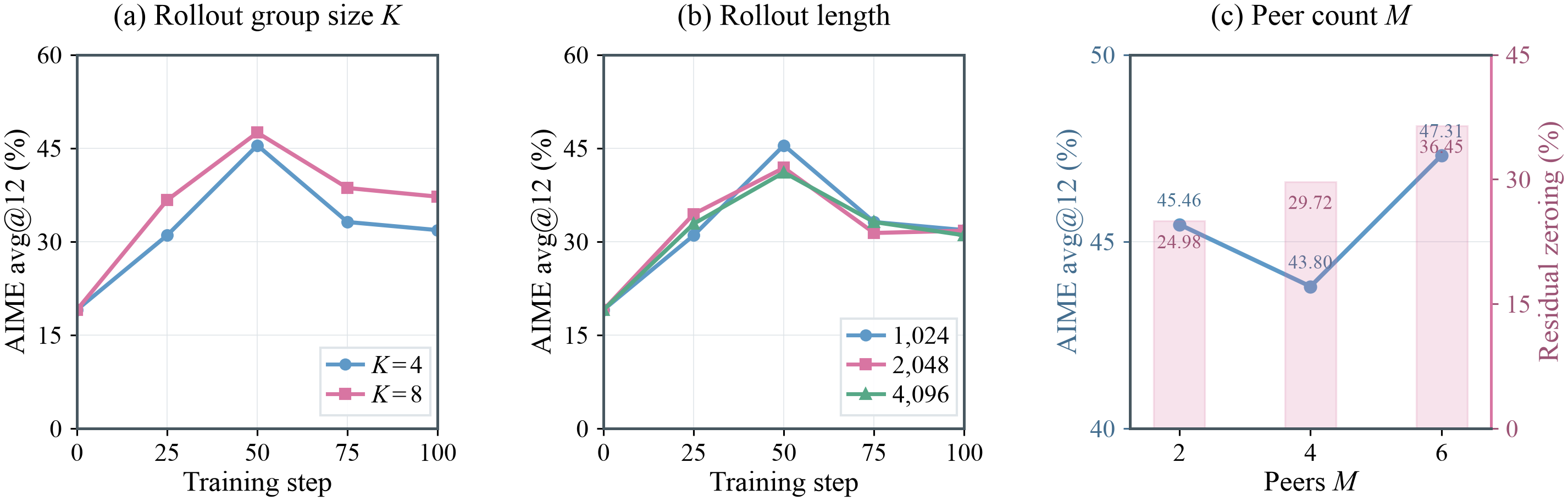}
  \vspace{-0.8em}
  \caption{Sampling sensitivity and peer-count analysis on Qwen3-4B. Panels (a)--(b) report mean AIME24--26 avg@12 throughout training while varying rollout group size and rollout length. Panel (c) jointly reports performance at the main-table checkpoint and the training-average residual zeroing rate induced by $M\in\{2,4,6\}$ peers, with two scored targets in every setting.}
  \label{fig:sensitivity_panels}
  \vspace{-0.5em}
\end{figure}

Figure~\ref{fig:sensitivity_panels} uses Qwen3-4B, the same optimizer schedule and decoding configuration, and mean AIME24--26 avg@12. Panels (a)--(b) track performance throughout training; panel (c) reports performance at the main-table checkpoint.

\paragraph{Sampling budgets.}
Panel (a) varies the total number $K$ of on-policy trajectories while fixing $M=2$; all remaining $K-2$ trajectories are scored targets. At the main-table checkpoint, $K=4$ achieves 45.46 mean avg@12 with half the rollout count of $K=8$, which achieves 47.50. Panel (b) varies rollout length at $K=4,M=2$. The teacher receives full sampled peer trajectories up to the generation limit, but only the first 1,024 target tokens are scored. Cross-problem controls are drawn from the same batch and follow the same generation budget. The 1,024-token setting reaches 45.46, compared with 41.85 and 41.11 for 2,048- and 4,096-token rollouts. These results support the default $K=4$ and 1,024-token generation budget.

\paragraph{Peer count and residual zeroing.}
Panel (c) fixes two scored target trajectories and varies $M\in\{2,4,6\}$, so each group contains $K=M+2$ rollouts. The cross-problem control is reused from another group and excluded from $M$. Equation~\eqref{eq:intersection} retains a coordinate only when all peer residuals have the same nonzero sign, taking the smallest magnitude. The residual zeroing rate counts token--vocabulary coordinates zeroed by this sign test, divided by all scored token--vocabulary coordinates. Numerators and denominators are aggregated globally across training before taking their ratio.

As $M$ increases from 2 to 4 and 6, the zeroing rate rises from 24.98\% to 29.72\% and 36.45\%. The default $M=2$ reaches 45.46 mean avg@12, while $M=4$ and $M=6$ reach 43.80 and 47.31, respectively. Two peers already provide effective guidance, while six peers offer a higher score at additional sampling and scoring cost.

\subsection{Peer Correctness Composition}
\label{app:peer_correctness_composition}

Table~\ref{tab:peer_correctness_composition} examines whether peers with incorrect final answers can provide useful supervision. Determining peer correctness requires an extractable final \texttt{\textbackslash boxed\{\}} answer, which is normalized and checked against the dataset reference. We therefore extend the rollout budget to obtain more completed answers and reduce truncation. In a single offline test on the OPSD training dataset, using a batch of 512 problems with $K=4$ rollouts per problem, the rollout truncation rate was 43.75\% at a 1,024-token limit and 3.03\% at a 4,096-token limit. Accordingly, all three conditions use 4,096-token rollouts while retaining the 1,024-token scoring limit.

We set $M=2$ and evaluate at the main-table checkpoint, with 1,600 groups accepted through dynamic sampling for each condition: two correct peers, one correct and one incorrect peer, or two incorrect peers. The dataset answer is used only to populate these diagnostic batches; the \methodabbr{} target, number of distillation targets, optimization schedule, and evaluation protocol remain fixed.

\begin{table}[!h]
\centering
\caption{Peer-correctness composition analysis on Qwen3-4B at the main-table checkpoint. Every condition uses 4,096-token rollouts, $M=2$ peers, and 1,600 accepted training groups. Reference answers are used only to determine pair composition. All AIME scores use avg@12, and Average is their arithmetic mean.}
\label{tab:peer_correctness_composition}
\footnotesize
\setlength{\tabcolsep}{5pt}
\renewcommand{\arraystretch}{1.0}
\begin{tabular*}{\linewidth}{@{\extracolsep{\fill}}lcccc@{}}
\toprule
\multicolumn{5}{l}{\textit{Evaluation at the main-table checkpoint}} \\
\textbf{Peer pair} & \textbf{AIME24} & \textbf{AIME25} & \textbf{AIME26} & \textbf{Average} \\
\midrule
Both correct & 40.00 & 38.61 & 38.61 & 39.07 \\
One correct, one incorrect & 45.83 & 37.78 & 38.89 & 40.83 \\
Both incorrect & 49.44 & 37.22 & 43.61 & 43.43 \\
\bottomrule
\end{tabular*}

\vspace{4pt}
\begin{tabular*}{\linewidth}{@{\extracolsep{\fill}}lrrr@{}}
\toprule
\multicolumn{4}{l}{\textit{Observed sampling and training volume}} \\
\textbf{Peer pair} & \textbf{Candidate groups} & \textbf{Accepted rollout tokens} & \textbf{Loss tokens} \\
\midrule
Both correct & 3,088 & 7,493,105 & 2,514,815 \\
One correct, one incorrect & 8,560 & 10,593,270 & 2,944,013 \\
Both incorrect & 9,360 & 10,343,735 & 2,896,163 \\
\bottomrule
\end{tabular*}

\vspace{2pt}
\parbox{\linewidth}{\scriptsize \textit{Note.} Accepted rollout tokens count all four trajectories within selected groups, each capped at 4,096 tokens. Loss tokens count non-padding tokens from the first 1,024 tokens of the two scored trajectories. Lengths of rejected candidates were not logged, so accepted rollout tokens do not measure total sampling cost.}
\end{table}

At the main-table checkpoint, mean AIME24--26 avg@12 rises from the base model's 19.07 to 40.83 with mixed pairs and 43.43 with two incorrect peers. Thus, peers with incorrect final answers can supply effective supervision, supporting the use of local distributional guidance from unverified solutions. The conditions use different selected problem groups and sampling and training volumes (Table~\ref{tab:peer_correctness_composition}); comparisons across compositions are therefore descriptive.

\FloatBarrier
\subsection{OPSD Scaling Control}
\label{app:opsd_scaling}

Table~\ref{tab:opsd_scaling_control} evaluates the target-batch adjustment to OPSD$^\ddagger$ at the main-table checkpoint. Original OPSD samples and scores one trajectory per problem ($K=1$), whereas the matched variant samples two independent trajectories and scores both ($K=2$). Thus, the comparison changes the number of on-policy targets without introducing peer aggregation or discarding additional rollouts. 
\begin{table}[!h]
\centering
\caption{OPSD target-batch scaling across model sizes (avg@12, \%; higher is better). $K=1$ is the original configuration and $K=2$ is the target-count-matched variant used in the main comparison. }
\label{tab:opsd_scaling_control}
\small
\setlength{\tabcolsep}{6pt}
\renewcommand{\arraystretch}{1.08}
\begin{tabular}{lccccc}
\toprule
\textbf{Model} & $\boldsymbol{K}$ & \textbf{Targets/update} & \textbf{AIME24} & \textbf{AIME25} & \textbf{AIME26} \\
\midrule
Qwen3-1.7B & 1 & 32 & 13.33 & 8.33 & 12.50 \\
            & 2 & 64 & 14.17 & 11.39 & 10.56 \\
\midrule
Qwen3-4B   & 1 & 32 & 23.61 & 24.44 & 24.44 \\
            & 2 & 64 & 22.50 & 26.39 & 25.00 \\
\midrule
Qwen3-8B   & 1 & 32 & 42.22 & 32.22 & 33.89 \\
            & 2 & 64 & 40.56 & 31.94 & 27.50 \\
\bottomrule
\end{tabular}
\end{table}

\FloatBarrier

\end{document}